\pdfoutput=1

\documentclass[11pt]{article}
\usepackage{graphicx}
\usepackage{subcaption}

\usepackage{acl}
\usepackage{times}
\usepackage{latexsym}
\usepackage{multirow}
\usepackage{dingbat}
\usepackage{amsmath}
\usepackage{tcolorbox}
\usepackage{enumitem}
\usepackage{calc}
\usepackage{tikz}
\usetikzlibrary{shapes.geometric, arrows}

\usepackage[T1]{fontenc}

\usepackage[utf8]{inputenc}

\usepackage{microtype}

\usepackage{inconsolata}

\usepackage{graphicx}

\title{In Defense of RAG in the Era of Long-Context Language Models}

\author{Tan Yu \\
 NVIDIA \\
 Santa Clara, California\\
 United States \\
  \texttt{tayu@nvidia.com} \\\And
  Anbang Xu \\
NVIDIA \\
Santa Clara, California \\
United States \\
  \texttt{anbangx@nvidia.com} \\\And
  Rama Akkiraju \\
NVIDIA \\
Santa Clara, California \\
United States \\
  \texttt{rakkiraju@nvidia.com} }

\begin{document}
\maketitle
\begin{abstract}

Overcoming the limited context limitations in early-generation LLMs, retrieval-augmented generation (RAG) has been a reliable solution for context-based answer generation in the past. Recently, the emergence of long-context LLMs allows the models to incorporate much longer text sequences, making RAG less attractive. Recent studies show that long-context LLMs significantly outperform RAG in 
long-context applications.  Unlike the existing works favoring the long-context LLM over RAG, we argue that the extremely long context in LLMs suffers from a diminished focus on relevant information and leads to potential degradation in answer quality.  This paper revisits the  RAG in long-context answer generation. We propose an order-preserve retrieval-augmented generation (OP-RAG) mechanism, which significantly improves the performance of RAG for long-context question-answer applications. With OP-RAG, as the number of retrieved chunks increases, the answer quality initially rises, and then declines, forming an inverted U-shaped curve. There exist sweet points where OP-RAG could achieve  higher answer quality with much less tokens than long-context LLM taking the whole context as input. Extensive experiments on public benchmark demonstrate the superiority of our OP-RAG.


\end{abstract}

\section{Introduction}

Due to the limited context window length (\emph{eg}, 4096) of early-generation large language models (LLMs),  retrieval augmented generation (RAG)~\cite{guu2020retrieval,lewis2020retrieval} is an indispensable choice to handle a large-scale context corpus. Since the answer quality is heavily dependent on the performance of the retrieval model, a lot of efforts are devoted to improving the retrieval recall/precision when designing the RAG system. 

\begin{figure}[t!]
    \centering
    \begin{subfigure}{0.54\textwidth}
        \centering
        \includegraphics[width=\textwidth]{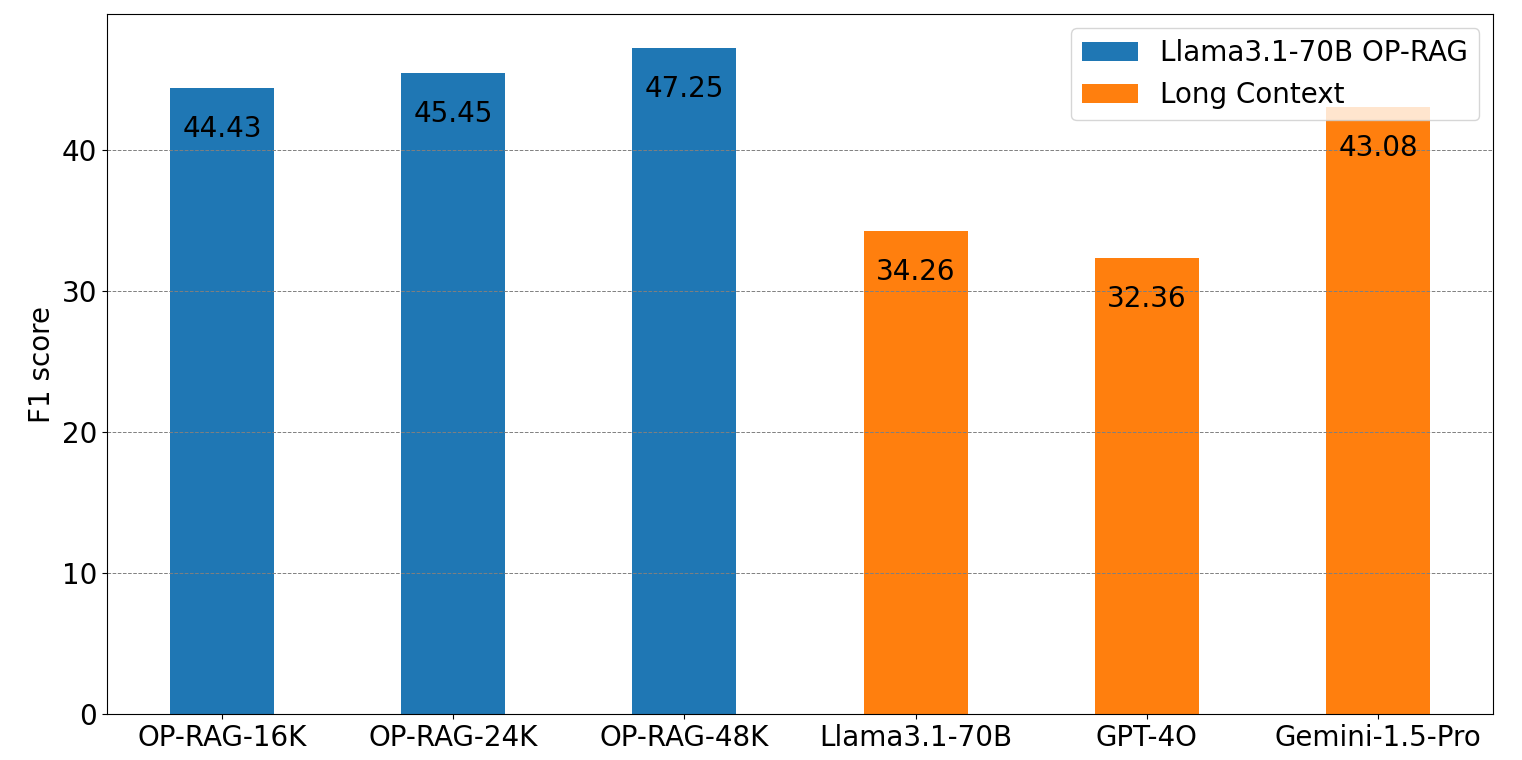}
        \caption{F1 score.}
        \label{fig:figure1}
    \end{subfigure}
    \hfill
    \begin{subfigure}{0.54\textwidth}
        \centering
        \includegraphics[width=\textwidth]{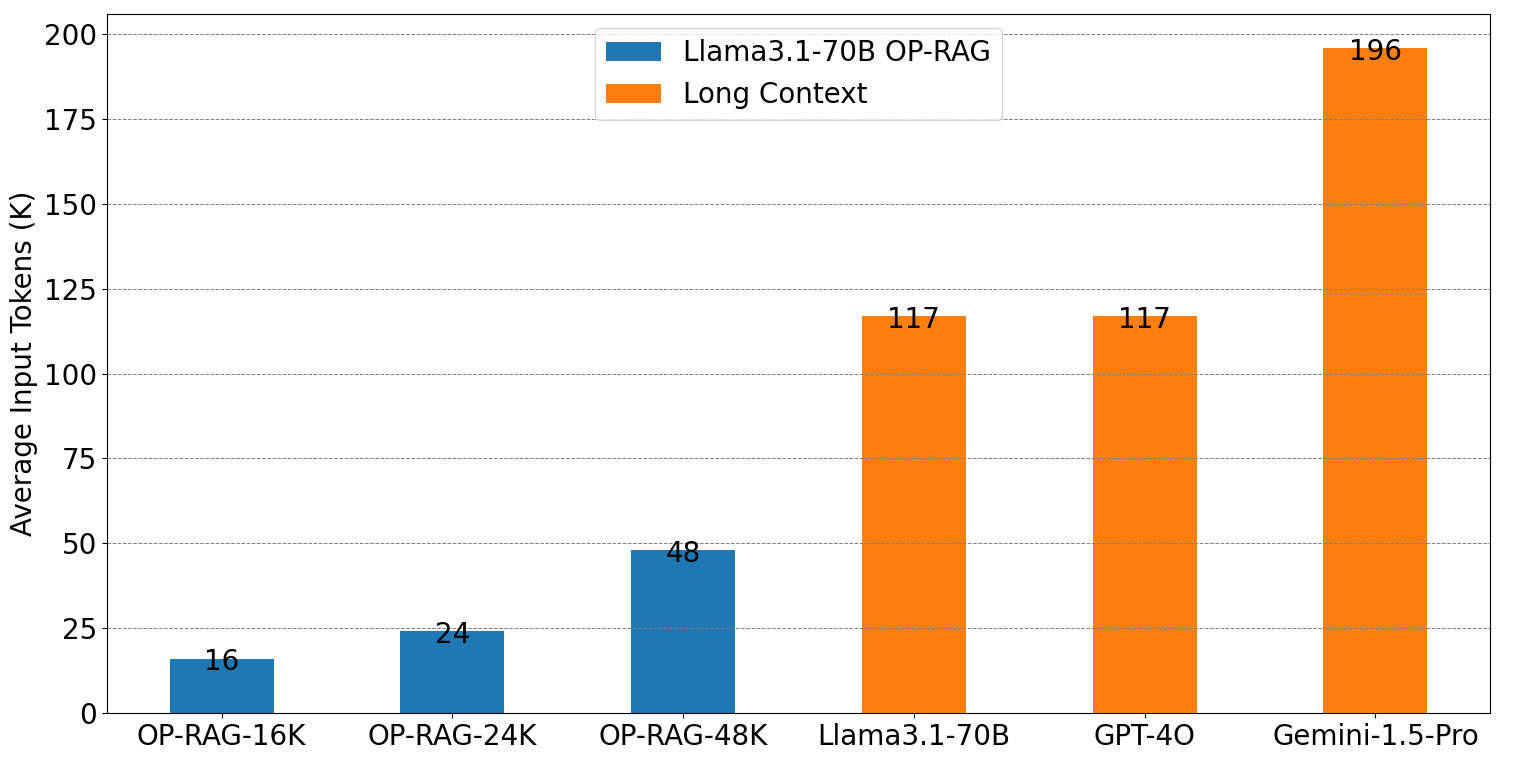}
        \caption{Input token count.}
        \label{fig:figure2}
    \end{subfigure}
    
    \caption{Comparisons between the proposed order-preserve retrieval-augmented generation (OP-RAG) and approaches using long-context LLMs without RAG on En.QA dataset of $\infty$Bench. Our OP-RAG uses Llama3.1-70B as generator, which significantly outperforms its counterpart using Llama3.1-70B without RAG. }
    \label{fig:combined_figures}
\end{figure}

Recently, the state-of-art LLMs support much longer context windows. For example, GPT-4O~\cite{openai2023gpt}, Claudi-3.5~\cite{Claude2024gpt}, Llama3.1~\cite{llama31modelcard}, Phi-3~\cite{abdin2024phi}, and Mistral-Large2~\cite{mistrallarge2}  all support 128-K context. Gemini-1.5-pro even supports a 1M context window.   The recent emergence of long-context LLMs naturally leads to the question: is RAG necessary in the age of long-context LLMs? \citet{li2024retrieval}
recently systematically compares RAG with long-context (LC) LLMs (w/o RAG) and demonstrates that LC LLMs consistently outperform RAG in terms of answer quality.

In this work, we re-examine the effectiveness of RAG in long-context answer generation.
We observe that the order of retrieved chunks in the context of LLM is vital for the answer quality.  Different from traditional RAG which places the retrieved chunks in a relevance-descending order, we propose to preserve the order of retrieved chunks in the original text. Our experiments show that the proposed order-preserving mechanism significantly improves the answer quality of RAG.

Meanwhile, using the proposed order-preserve RAG, as the number of retrieved chunks increases, the answer quality initially rises and then declines. This is because, with more retrieved chunks, the model has access to more potentially relevant information, which improves the chances of retrieving the correct context needed to generate a high-quality answer. However, as more chunks are retrieved, the likelihood of introducing irrelevant or distracting information also increases. This excess information can confuse the model, leading to a decline in answer quality. The trade-off, therefore, is between improving recall by retrieving more context and maintaining precision by limiting distractions. The optimal point is where the balance between relevant and irrelevant information maximizes the quality of the answer. Beyond this point, the introduction of too much irrelevant information degrades the model’s performance. It explains the inferior performance of the approach taking the whole long context as the input of LLM.



Different from the conclusion from \citet{li2024retrieval},  with the proposed order-preserving mechanism, RAG achieves higher answer quality compared with its counterparts that rely solely on Long-Context LLMs.  
As shown in Figure~\ref{fig:figure1}, 
On En.QA dataset of $\infty$Bench~\cite{zhang2024inftybench}, using only $16$K retrieved tokens, we achieve $44.43$ F1 score with Llama3.1-70B. In contrast, without RAG, Llama3.1-70B making full use of $128$K context only achieves $34.32$ F1 score, GPT-4O achieves only $32.36$ F1 score and Gemini-1.5-Pro obtains only $43.08$ F1 score as evaluated by \citet{li2024retrieval}. That is,  RAG could achieve a higher F1 score even with a significant reduction on input length.

\section{Related Work}

\textbf{Retrieval-augmented generation.} By incorporating the external knowledge as context, retrieval-augmented generation (RAG)~\cite{guu2020retrieval,lewis2020retrieval,mialon2023augmented} allows language model to access up-to-date and specific information, reducing hallucinations and improving factual accuracy. Before the era of long-context LLMs, RAG  is a promising solution to overcoming  the limitation of short context window. 

\noindent
\textbf{Long-context LLM.} 
To support the long sequence of language models, many efforts have been devoted to improving the computing efficiency of self-attention~\cite{choromanski2020rethinking,zaheer2020big,tay2020efficient,dao2022flashattention,dao2023flashattention2} and boosting extensibility of positional encoding~\cite{press2021train,sun2022length, chen2023extending}. Recently, the flagship LLMs such as GPT-4O~\cite{openai2023gpt}, Gemini-1.5-Pro~\cite{reid2024gemini}, Claudi-3.5~\cite{Claude2024gpt},  Grok-2~\cite{grok2024}, and Llama3.1~\cite{Llama2024} have supported extremely large context. With the existence of long-context LLMs, RAG is no longer a indispensable module for long-context question-answering task.  Recently, \citet{li2024retrieval} concludes that using long-context without RAG could significantly outperforms RAG. Different from the conclusion from~\cite{li2024retrieval}, in this work, we demonstrate the proposed order-preserve RAG could beat the long-context LLMs without RAG. 



\begin{figure}
    \centering
    \includegraphics[width=1\linewidth]{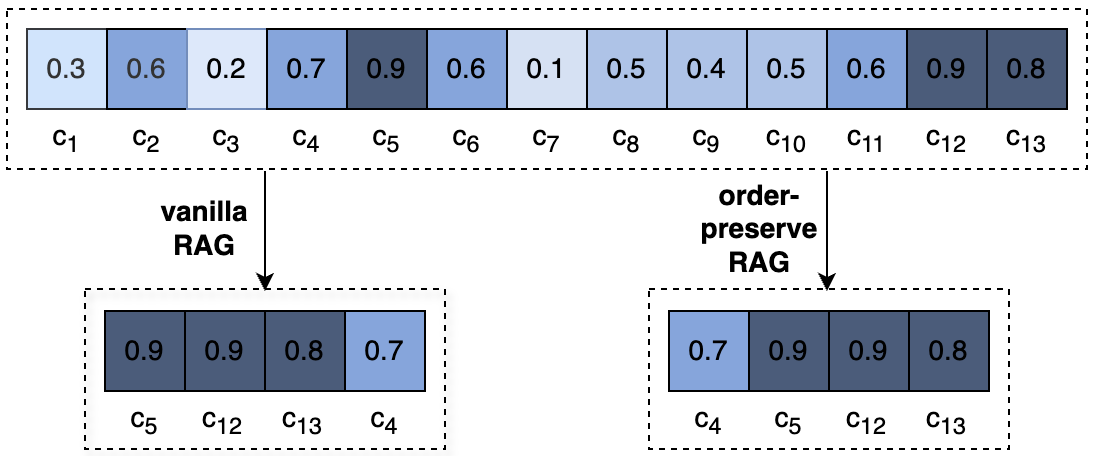}
    \caption{Vanilla RAG versus the proposed order-preserve the RAG. As shown in the example, a long document is cropped into $13$ chunks, $\{c_i\}_{i=1}^{13}$. The similarity score is appended to each chunk.  We retrieve top 4 chunks with the highest similarity scores.
    Vanilla RAG places the chunks in a score-descending order, whereas the proposed order-preserve RAG places the chunks based on the order in the original document.}
    \label{fig:oprag}
\end{figure} 
 
\section{Order-Preserve RAG}
 \begin{figure*}[htp!]
    \centering
    \begin{subfigure}{0.45\textwidth}
        \centering
        \includegraphics[width=\textwidth]{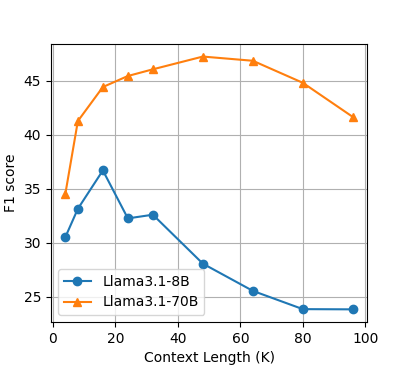}
        \caption{EN.QA}
        \label{fig:figure1}
    \end{subfigure}
    \begin{subfigure}{0.45\textwidth}
        \centering
        \includegraphics[width=\textwidth]{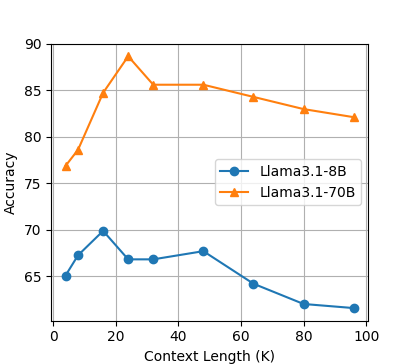}
        \caption{EN.MC}
        \label{fig:figure2}
    \end{subfigure}
    
    \caption{The influence of context length on the performance of RAG. The evaluations are conducted on En.QA and EN.MC datasets of $\infty$Bench. }
    \label{fig:context_length}
\end{figure*}

Let us denote the long textual context, \emph{e.g.}, a long document, by $d$.
We split $d$ into $N$ chunks sequentially and uniformly, $\{c_i\}_{i=1}^N$. 
The index $i$ implies the sequential order of the chunk $c_i$ in $d$. 
That is, $c_{i-1}$ denotes the chunk before $c_i$ whereas $c_{i+1}$ denotes the chunk right after $c_i$. Given a query $q$, we obtain the relevance score of the chunk $c_i$ by computing cosine similarity between the embedding of $q$ and that of $c_i$:

\begin{equation}
    s_i = \mathrm{cos}(\mathrm{emb}(q), \mathrm{emb}(c_i)),
\end{equation}
where $\mathrm{cos}(\cdot,\cdot)$ denotes the cosine similarity function and $\mathrm{emb}(\cdot)$ denotes the embedding function.

We retrieve the top k chunks with the highest similarity scores with the query and 
denote the indices of top k chunks by $\mathcal{J} = \{j_i\}_{i=1}^k$. We preserve the  order of chunks in the original long context $d$, that is, we constrain

\begin{equation}
    j_l > j_m \iff l > m.
\end{equation}
Figure~\ref{fig:oprag}  visualizes the difference between the vanilla RAG and the proposed order-preserve RAG. Different from vanilla RAG placing the chunks in the order of similarity descending, the proposed order-preserve RAG keep the order of chunks in the original document.

\section{Experiments}

\subsection{Datasets.}
We conduct experiments on EN.QA and EN.MC datasets of $\infty$Bench~\cite{zhang2024inftybench} benchmark, specially designed for long-context QA evaluation. To be specific, En.QA consists of 351 human-annotated question-answer pairs. On average, the long context in En.QA contains 150,374 words.
We use F1-score as metric for evaluation on En.QA. 
EN.MC consists of 224 question-answer pairs, which are annotated similarly to En.QA, but each question is provided with four answer choices. On average, the long context in En.MC contains 142,622 words.
We use accuracy as metric for evaluation on En.QA.  We notice there is another benchmark termed LongBench~\cite{bai2023longbench}. Nevertheless, the average context length of LongBench is below 20K words, which is not long enough to evaluate the recent long-context LLMs supporting 128K-token window size.

\begin{figure*}[htp]
    \centering
    \begin{subfigure}{0.48\textwidth}
        \centering
        \includegraphics[width=\textwidth]{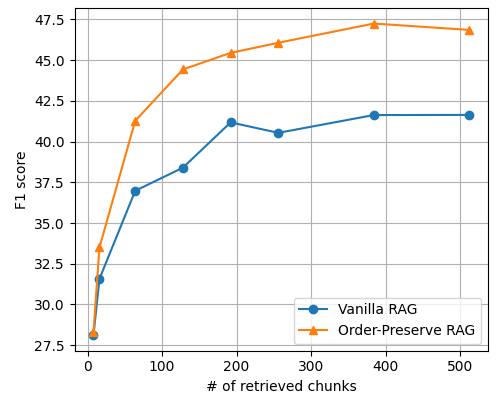}
        \caption{EN.QA}
        \label{fig:figure1}
    \end{subfigure}
    \begin{subfigure}{0.48\textwidth}
        \centering
        \includegraphics[width=\textwidth]{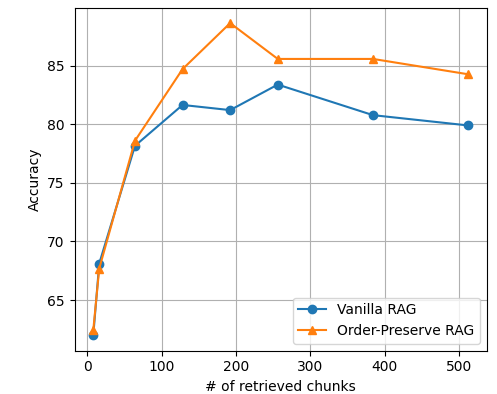}
        \caption{EN.MC}
        \label{fig:figure2}
    \end{subfigure}
    
    \caption{Comparisons between the proposed order-preserve RAG and vanilla RAG. The evaluations are conducted on En.QA and EN.MC datasets of $\infty$Bench, using Llama3.1-70B model. }
    \label{fig:comp}
\end{figure*}

\subsection{Implementation details.}
We set the chunk size as $128$ tokens on all datasets. Chunks are non-overlapped. 
We use BGE-large-en-v1.5~\cite{bge_embedding} to extract the embedding of queries and chunks, by default.

\subsection{Ablation Study}

\textbf{The influence of context length.} We evaluate the influence of the context length on the performance of the proposed order-preserve RAG. Since each chunk contains $128$ tokens, the context length is $128m$, where $m$ is the number of the retrieved chunks as the context for generating the answer. As shown in Figure~\ref{fig:context_length}, as the context length increases, the performance initially increases. This is because more context might have a greater chance of covering the relevant chunk. Nevertheless, as the 
context length further increases, the answer quality drops since more irrelevant chunks are used as distractions. To be specific, Llama3.1-8B model achieves the performance peak when the context length is 16K on both EN.QA dataset and EN.MC dataset, whereas the best performance of  Llama3.1-70B model is achieved at $48$K on EN.QA and $32$K on EN.MC. 
The fact that the peak point of Llama3.1-70B comes later than Llama3.1-8B model might be because the larger-scale model has a stronger capability to distinguish the relevant chunks from irrelevant distractions.

\noindent \textbf{Order-preserve RAG versus vanilla RAG.}
As shown in Figure~\ref{fig:comp}, when the number of retrieved chunks are small (\emph{e.g}, 8), the advantage of the proposed order-preserve RAG over vanilla RAG is not considerably.  In contrast, when the number of retrieved chunks is large, our order-preserve RAG significantly outperforms vanilla RAG. To be specific, on EN.QA dataset, when the number of retrieved chunk is $128$, vanilla RAG only achieves $38.40$ F1-score whereas our  order-preserve RAG  achieves $44.43$ F1-score. On EN.MC dataset, retrieving $192$ chunks, vanialla RAG only achieves $81.22$ accuracy whereas our order-preserve RAG obtains $88.65$ accuracy.

\subsection{Main Results}
We compare the proposed order-preserve RAG with two types of baselines. The first category of approaches uses the long-context LLM without RAG.  As shown in Table~\ref{tab:main}, without RAG, LLM takes a huge number of tokens as input, which is inefficient and costly. In contrast, the proposed order-preserve RAG not only significantly reduces the number of tokens, but also significantly improves the answer quality. For instance, using Llama3.1-70B model, the approach without RAG only achieves a $34.26$ F1 score on EN.QA with an average of 117K tokens as input. In contrast, our OP-RAG with 48K tokens as input attains a $47.25$ F1 score. The second category of baselines takes the SELF-ROUTE mechanism~\cite{li2024retrieval}, which routes queries to RAG or long-context LLM based on the model self-reflection. As shown in Table~\ref{tab:main}, ours significantly outperforms than using much fewer tokens in the input of LLMs.

\begin{table}[]
\small
\begin{tabular}{c|cc|cc}
\hline

\multirow{2}{*}{Method} & \multicolumn{2}{c|}{EN.QA}                & \multicolumn{2}{c}{EN.MC}                \\ \cline{2-5} 
                        & \multicolumn{1}{l|}{F1 Score} & Tokens & \multicolumn{1}{l|}{Acc.} & Tokens \\ \hline \hline
         
 \multicolumn{5}{c}{ Long-context LLM w/o RAG }                                 \\
                        \hline \hline 
Llama3.1-70B            & \multicolumn{1}{c|}{$34.26$}         &   117K        & \multicolumn{1}{c|}{$71.62$}         &    117K       \\ 
GPT-4O                    & \multicolumn{1}{c|}{$32.36$}         &   117K        & \multicolumn{1}{c|}{$78.42$}         &    117K           \\ 
Gemini-1.5-Pro          & \multicolumn{1}{c|}{$43.08$}         &   196K      & \multicolumn{1}{c|}{$85.57$}         &      188K     \\ \hline  \hline
 \multicolumn{5}{c}{SELF-ROUTE~\cite{li2024retrieval} }        \\ \hline \hline

 GPT-4O                    & \multicolumn{1}{c|}{$34.95$}         &   85K        & \multicolumn{1}{c|}{$77.29$}         &    62K           \\ 
Gemini-1.5-Pro          & \multicolumn{1}{c|}{$37.51$}         &   83K      & \multicolumn{1}{c|}{$76.86$}         &      62K     \\ \hline  \hline
 \multicolumn{5}{c}{Llama3.1-70B order-preserve RAG (ours) }     \\
                        \hline \hline 
OP-RAG-16K              & \multicolumn{1}{c|}{$44.43$}         &     16K      & \multicolumn{1}{c|}{$84.72$}         &    16K       \\ 
OP-RAG-24K              & \multicolumn{1}{c|}{$45.45$}         &       24K    & \multicolumn{1}{c|}{$\mathbf{88.65}$}         &     24K      \\ 
OP-RAG-48K              & \multicolumn{1}{c|}{$\mathbf{47.25}$}         &       48K    & \multicolumn{1}{c|}{$85.59$}         &      48K     \\ \hline
\end{tabular}
\caption{Comparisons among the long-context LLM without RAG, SELF-ROUTE mechanism~\cite{li2024retrieval}  and the proposed order-preserve (OP) RAG.}
\label{tab:main}
\end{table}

\section{Conclusion}
In this paper, we have revisited the role of retrieval-augmented generation (RAG) in the era of long-context language models (LLMs). While recent trends have favored long-context LLMs over RAG for their ability to incorporate extensive text sequences, our research challenges this perspective. We argue that extremely long contexts in LLMs can lead to a diminished focus on relevant information, potentially degrading answer quality in question-answering tasks.
To address this issue, we proposed the order-preserve retrieval-augmented generation (OP-RAG) mechanism. Our extensive experiments on public benchmarks have demonstrated that OP-RAG significantly improves the performance of RAG for long-context question-answer applications.  OP-RAG's superior performance suggests that efficient retrieval and focused context utilization can outperform the brute-force approach of processing extremely long contexts.
\bibliography{custom}

\appendix

\end{document}